\documentclass[sigconf]{acmart}

\usepackage{booktabs,multirow,threeparttable}

\usepackage{amsmath}
\usepackage{amssymb}
\usepackage{bbm}
\usepackage{mathtools}

\usepackage{url}

\usepackage[edges]{forest}
\usepackage{tikz}
\usepackage{pgfplots}
\usepackage{tabularx}
\usepackage{float} 
\usepackage{tabulary}
\usepackage{subfig}

\def\BibTeX{{\rm B\kern-.05em{\sc i\kern-.025em b}\kern-.08emT\kern-.1667em\lower.7ex\hbox{E}\kern-.125emX}}
    
\copyrightyear{2019}
\acmYear{2019}
\setcopyright{acmlicensed}
\acmConference[KDD, DsHealth 2019]{2019 KDD workshop on Applied data science in Healthcare: bridging the gap between data and knowledge}{Anchorage, AK}
\begin{document}

%
% The "title" command has an optional parameter, allowing the author to define a "short title" to be used in page headers.
\title{Activity2Vec: Learning ADL Embeddings from Sensor Data with a Sequence-to-Sequence Model}

%
% The "author" command and its associated commands are used to define the authors and their affiliations.
% Of note is the shared affiliation of the first two authors, and the "authornote" and "authornotemark" commands
% used to denote shared contribution to the research.

\author{Alireza Ghods}
\affiliation{%
  \institution{Washington State University}}
\email{alireza.ghods@wsu.edu}

\author{Diane J. Cook}
\affiliation{%
  \institution{Washington State University}}
\email{djcook@wsu.edu}

%
% The abstract is a short summary of the work to be presented in the article.
\begin{abstract}
Recognizing activities of daily living (ADLs) plays an essential role in analyzing human health and behavior. The widespread availability of sensors implanted in homes, smartphones, and smart watches have engendered collection of big datasets that reflect human behavior. To obtain a machine learning model based on these data,researchers have developed multiple feature extraction methods. In this study, we investigate a method for automatically extracting universal and meaningful features that are applicable across similar time series-based learning tasks such as activity recognition and fall detection. We propose creating a sequence-to-sequence (seq2seq) model to perform this feature learning. Beside avoiding feature engineering, the meaningful features learned by the seq2seq model can also be utilized for semi-supervised learning. We evaluate both of these benefits on datasets collected from wearable and ambient sensors.
\end{abstract}

 \begin{CCSXML}
<ccs2012>
<concept>
<concept_id>10003752.10010070.10010071</concept_id>
<concept_desc>Theory of computation~Machine learning theory</concept_desc>
<concept_significance>500</concept_significance>
</concept>
<concept>
<concept_id>10003752.10010070.10010071.10010074</concept_id>
<concept_desc>Theory of computation~Unsupervised learning and clustering</concept_desc>
<concept_significance>500</concept_significance>
</concept>
<concept>
<concept_id>10003752.10010070.10010071.10010083</concept_id>
<concept_desc>Theory of computation~Models of learning</concept_desc>
<concept_significance>500</concept_significance>
</concept>
<concept>
<concept_id>10010147.10010257.10010293.10010294</concept_id>
<concept_desc>Computing methodologies~Neural networks</concept_desc>
<concept_significance>500</concept_significance>
</concept>
</ccs2012>
\end{CCSXML}

\ccsdesc[500]{Theory of computation~Machine learning theory}
\ccsdesc[500]{Theory of computation~Models of learning}
\ccsdesc[500]{Theory of computation~Unsupervised learning and clustering}
\ccsdesc[500]{Computing methodologies~Neural networks}

%
% Keywords. The author(s) should pick words that accurately describe the work being
% presented. Separate the keywords with commas.
\keywords{activity embeddings, sequence-to-sequence learning, recurrent neural networks, unsupervised learning.}

%
% This command processes the author and affiliation and title information and builds
% the first part of the formatted document.
\maketitle

\section{Introduction}

Data collected from an ambient sensor environment or body-worn sensors offer valuable insights on human behavior and the relationship between behavior and human health. These are time series data, where each data point is paired with a time-stamp. Similar to other time series problems, the processing pipeline has three stages: segmentation of the input stream, feature extraction, and classification. The classification here could be activity recognition, health-based diagnosis of the individual generating sensor data, or other clinical tasks. The key to creating an effective classifier mostly relies on defining high-quality features. The two major approaches that have been employed to extract features are hand-crafting of features and feature learning methods. Hand-crafted features require expert knowledge about the data and are not generalizable across different problems \cite{clark2017review}. Feature learning approaches exploit the power of deep networks to transform raw data through a non-linear transformation. In contrast to feature engineering, feature learning methods do not need experts and can be applied across different problems. Existing feature learning approaches can be categorized into two general groups: supervised and unsupervised. Whereas supervised models require a large amount of labeled data, unsupervised models can learn features from unlabeled data.

Autoencoding represents a prominent unsupervised approach which extracts features for training a machine learning model. An autoencoder is a specific type of neural network which copies its inputs, by way of representation transformation, to its outputs. An autoencoder consists of two parts: an encoder which maps inputs to a new representation $f(x)$, and a decoder which takes the new representation and attempts to reconstruct the original input, $g(f(x))$.

In the context of data collected while an individual performs Activities of Daily Living, or ADL sensor data, employment of autoencoders has been investigated as being a general approach for feature extraction.  Unger et al. \cite{unger2016towards}  increase the recommendation power of their context-aware recommender system by extracting features from a smartphone with various sensors such as GPS, accelerometer, and gyroscope by applying autoencoders. Munoz-Organero et al. \cite{munoz2017time} applied autoencoding to extract features from data collected via a hand-worn accelerometer for activity recognition. Khan et al.  \cite{khan2017detecting} proposed an ensemble of autoencoders. Here, each autoencoder processes a different channel of accelerometer/gyroscope data. The result represents a concatenation of each autoencoder and is input to a fall detection model. Wang et al. \cite{wang2016human} employed a denoising autoencoder to learn feature representations for smart home data.

Even with the promising power of autoencoders for feature extraction on ADL data, a multilayer perceptron autoencoder, such as the ones applied in the above studies, does not take into account temporal dependencies. Time series data is not i.i.d., and ignoring the inherent feature inter-dependencies can lead to skewed results because the reading at time $t$ is heavily correlated with the readings at times $t-1$ and $t+1$. Similarly, data collected from ambient sensors located throughout a smart home will exhibit spatial relationships that need to be captured.

Yet another shortcoming of the multilayer perceptron autoencoder is its inability to handle time series input data of different-sized time windows. If a fixed-time input is required, then the data must first be segmented before it can be modeled and learned. Common segmentation methods typically produce data sequences that partition the data into either fixed-sized windows or dynamic-sized windows \cite{aminikhanghahi2019enhancing}. Therefore, autoencoders that process ADL-based sensor data must be able to handle dynamic input lengths. 

The goal of this study is to explore the feasibility of activity2vec framework, a seq2seq-type model, for extracting compact and abstract features from sensor data. We investigate the practicality of this method on two major sources of sensor data: ambient sensor data collected in smart homes and mobile sensor data collected from wearable devices. Additionally, we also examine the possibility of applying the proposed activity2vec framework to generate features for unseen classes.

\section{Sequence to Sequence Model}

\begin{figure}[t]
	\centering
	\begin{tikzpicture}
	    
	    \draw[fill=red,opacity=0.2,draw=black] (0,0) -- (0.5,0.0) -- (0.5,0.5) -- (0,0.5) -- (0,0) node [opacity=1,right,midway]{$x^1$};
	    \draw [-] (0.25,0.5) -- (0.25,1.1) [->] (0.25,1.1) -- (0.5,1.1);
	    \draw[fill=red,opacity=0.2,draw=black] (1.4,0) -- (1.9,0.0) -- (1.9,0.5) -- (1.4,0.5) -- (1.4,0) node [opacity=1,right,midway]{$x^2$};
	    \draw [-] (1.65,0.5) -- (1.65,1.1) [->] (1.65,1.1) -- (1.9,1.1);
	    \draw[fill=red,opacity=0.2,draw=black] (2.8,0) -- (3.3,0.0) -- (3.3,0.5) -- (2.8,0.5) -- (2.8,0) node [opacity=1,right,midway]{$x^t$};
	    \draw [-] (3.05,0.5) -- (3.05,1.1) [->] (3.05,1.1) -- (3.3,1.1);
	    \draw [->] (0.,1.25) -- (0.5,1.25)  node [opacity=1,above,midway]{$h_1$};
	    \draw[fill=black,opacity=0.2,draw=black] (0.5,1.) -- (1.4,1.) -- (1.4,1.5) -- (0.5,1.5) -- (0.5,1.)  node [opacity=1,right,midway]{RNN};
	    \draw [->] (1.4,1.25) -- (1.9,1.25) node [opacity=1,above,midway]{$h_2$};
	    \draw[fill=black,opacity=0.2,draw=black] (1.9,1.) -- (2.8,1.) -- (2.8,1.5) -- (1.9,1.5) -- (1.9,1.) node [opacity=1,right,midway]{RNN};
	    \draw [->] (3.29,1.25) -- (3.3,1.25);
	    \draw[fill=black,opacity=0.2,draw=black] (3.3,1.) -- (4.2,1.) -- (4.2,1.5) -- (3.3,1.5) -- (3.3,1.) node [opacity=1,right,midway]{RNN};
	    \draw [->] (4.2,1.25) -- (4.7,1.25);
	    \node at (3.0,1.25) {. . .};
	    \node at (3.1,1.5) {$h_t$};
	    \node at (2.5,2) {Encoder};
	    
	    \draw[fill=white,opacity=1.0,draw=black] (4.7,1.) -- (5.2,1.) -- (5.2,2.) -- (4.7,2.) -- (4.7,1.) node [right,midway] {z};
        \node [rotate=-90] at (4.9,0.5) {\scriptsize{Embedding}};
        \draw [->] (5.2,1.75) -- (5.7,1.75);
        \draw[fill=black,opacity=0.2,draw=black] (5.7,2.) -- (6.6,2.) -- (6.6,1.5) -- (5.7,1.5) -- (5.7,2.) node [opacity=1,right,midway]{RNN};
        \draw [->] (7.09,1.75) -- (7.1,1.75);
        \node at (6.8,1.75) {. . .};
        \draw[fill=black,opacity=0.2,draw=black] (7.1,2.) -- (8.,2.) -- (8.,1.5) -- (7.1,1.5) -- (7.1,2.) node [opacity=1,right,midway]{RNN};
        
        \draw [->] (6.15,2) -- (6.15,2.5);
	    \draw[fill=red,opacity=0.4,draw=black] (5.9,2.5) -- (6.4,2.5) -- (6.4,3.) -- (5.9,3) -- (5.9,2.5) node [opacity=1,right,midway]{$y^1$};
	    
	    \draw [->] (7.55,2) -- (7.55,2.5);
	    \draw[fill=red,opacity=0.4,draw=black] (7.3,2.5) -- (7.8,2.5) -- (7.8,3.) -- (7.3,3) -- (7.3,2.5) node [opacity=1,right,midway]{$y^t$};
        \node at (7,1) {Decoder};
	\end{tikzpicture}
	\caption{The structure of seq2seq model.}
    \label{figure:seq2seq}
\end{figure}
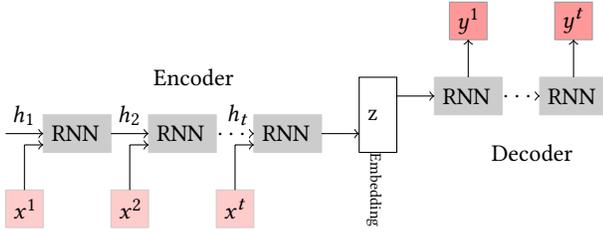

The seq2seq algorithm was introduced by Google in 2014 for translating a sentence in one language to another language (i.e., English to French) \cite{sutskever2014sequence}. Since that time, the seq2seq model has been deployed for voice capturing \cite{prabhavalkar2017comparison} and generating video descriptions \cite{venugopalan2015sequence}. The seq2seq model is an autoencoder where both the encoder and the decoder contain several recurrent units (these are typically LSTMs \cite{hochreiter1997long} or GRUs) as shown in Figure \ref{figure:seq2seq}. Given a sequence of inputs $x = (x^1, \cdots, x^t)$, the encoder learns a function $f$ that computes a hidden state $h_t$ at time step $t$, as follows:
\begin{equation}
    h_t = f(W^{(hh)}h_{t-1} + W^{(hx)}x^{t}),
\end{equation}
where $W^{(hh)}$ is the weight of connections between pairs of RNN units, and $W^{(hx)}$ is the weight of connections between input units and RNN units. The final hidden state, $z$, produced from the encoder aims to encapsulate the information for all input elements in order to help the decoder make accurate predictions. The final hidden state $z$ acts as the initial hidden state for the decoder. The decoder computes the hidden state $h_t$ as follows:
\begin{equation}
  h_t = f(W^{(hh)}h_{t-1}),
\end{equation}
and the output $y^t$ at time step $t$ is computed as follows:
\begin{equation}
    y_t = softmax(W^{hy}h_t),
\end{equation} 
where $W^{hy}$ is the weight vector from hidden units to output values.
%seq2seq figure

\subsection{Activity2Vec}

Inspired by the seq2seq model, we propose the activity2vec framework which maps a sequence of sensor readings to itself for learning a set of features from raw sensor data. Given a stream of sensor events $(x^1, x^2, \cdots, x^t)$, a fixed-length window or a dynamically-sized window can segment the data. Given a window of size $k$, data is segmented into subsequences $(s_1, s_2, \cdots, s_n)$, where $s_{1:k} = (x^1, \cdots, x^k)$. Throughout the training of activity2vec, the encoder takes the sequence $s_i$ as input and encodes it into a fixed-dimension vector representation $z_i$. Next, the decoder decodes $z_i$ to the original input $\hat{s}_i$. This model is trained by minimizing the reconstruction error $\sum_i ||s_i - \hat{s}_i||^2$, where $\hat{s}$ is the prediction of the model. The intuition behind this approach is that the decoder is unable to reconstruct the original feature $s_i$ if the encoded representation $z_i$ does not contain sufficient semantic information about that sequence.

We implemented the activity2vec model with Keras \cite{chollet2015keras}. The encoder is a single-layered bidirectional LSTM and the decoder is a single-layered unidirectional LSTM. By employing a bidirectional LSTM as the encoder, the activity2vec model is able to preserve information from both past and future. We also observe that imposing a $L1$ norm penalty on the encoder helps it to learn a better representation of $z_i$. The reason behind applying the $L1$ norm penalty is to penalize the less impactful features of the encoded representation $z_i$ in reconstruction of sequence $s_i$. During training, we added Gaussian noise to the input to make the activity2vec model indifferent to small changes. In our experiments, we trained the activity2vec model by applying the Adam \cite{kingma2014adam} optimizer for 1000 epochs. Once this is done, the trained encoder can be used to generate sensor embedding $z_i$ for each sequence $s_i$.

\section{Experiments}

We hypothesize that activity2vec can automatically extract features from sensor-based time series data. Furthermore, we anticipate that these automatically-learned features can facilitate comparable or superior representational performance for classification tasks that are valuable in clinical settings such as activity recognition and fall detection. Here we validate this hypothesis using collected sensor data.

\subsection{Dataset}

For our validation, we selected two available datasets. The first is the Human Activity Recognition Using Smartphones (HAR) dataset \cite{anguita2013public} collected from 30 volunteers in a lab performing six scripted different activities while wearing a smartphone on their wrist. The sensor signals (accelerometer and gyroscope) were sampled in fixed-width sliding windows of 2.56 seconds (128 readings were collected for each activity for each volunteer). We compare models learned from features generated by activity2vec with models learned directly from raw sensor data and from hand-crafted features. The 561 hand-crafted features were extracted by calculating statistical values such as mean, variance, minimum, and maximum from each reading. The HAR dataset has been randomly partitioned into two sets, where 70\% of the volunteers were selected for generating the training data and the remaining 30\% represents the test data. 

We also selected one of the CASAS smart homes (HH101) \cite{cook2013casas} datasets to measure the performance of activity2vec on data collected from ambient sensors. HH101 is a single-resident home equipped with passive infrared motion sensors in each room as well as binary sensors to detect movement of doors and cabinets. The data consists of four tuples <\textit{activation time, sensor, sensor status, activity label}>, (e.g., <\textit{2012-08-17 19:02:02.677811, D002, OPEN, Enter Home}>). Data were collected for two months while the resident performed normal daily routines, and were annotated by experts with twelve ground-truth activity labels to use for training activity classifiers. Data were segmented by moving a fixed-size window (thirty readings in length) with overlap. The 103 hand-crafted features were extracted by calculating statistical values such as the number of activated sensors, dominant sensor, start, end, and duration from each window. The HH101 dataset was partitioned into two sets of training and testing data by selecting four days of data as the training data and the following two days as the testing data, repeating this process for the entire two months of data.

\subsection{Evaluation Metric}

To evaluate the quality of features, we measure the predictive performance of a model learned from the activities. We also report intra-class similarities, and distances between learned classes. We select the Random Forest (RF) model as a benchmark classifier to asses that learned features by activity2vec can be utilized by classifier models. We implemented a RF model with 100 trees using scikit-learn \cite{scikit-learn}. 

In our experiments, intra-class similarity measures the similarity of features among samples of each class. We calculate the average Euclidean distance for all pairs of instances within each class. Finally, we evaluate the ability of the autoencoder to generate descriptive features for instances of a class which were not explicitly used for training.

\subsection{Results and Discussion}

We trained activity2vec models with different embedding sizes to establish the minimum embedding size that is needed to capture sufficient semantic information about the sensor data. We observe that increasing the embedding size does not always result in improved performance.

We trained the activity2vec model to encode the HAR and HH101 data to 128  features. To compare the capability of encoded features via acticvity2vec for building a classifier, we trained three RF models based on raw features, handcrafted features, and the activity2vec encoded features. We also normalized the handcrafted features for both datasets. As shown in Table \ref{table:HAR-f1score}, activity2vec achieves higher F1 scores on most classes except for two classes (sitting and standing) on HAR data. Before training the activity2vec model on HH101 data, we applied one-hot sensor encoding and added a time of day feature. Again, activity2vec features perform comparably with handcrafted features in many cases. However, the results are not consistently superior across all classes. Because the HH101 data were collected from a real environment, the number of samples for some classes such as bed to toilet and eat are very low. Another difficulty with HH101 data is coarse granularity of information provided by a limited number of installed sensors in the home. Therefore, some pairs of activities generate very similar sequences of sensor readings. For these reasons, activity2vec was unable to learn adequate representations compare to hand-crafted features for some classes, as shown in Table \ref{table:HH101-f1score}. Although the encoded features did not noticeably improve the accuracy of the classifier, our results demonstrate that the activity2vec framework offers comparable representational performance with hand-crafted features, without the time and expertise that is required to engineer such features. 

Next, we want to confirm that activity2vec can not only learn effective features for modeling predefined classes but can also encode appropriate features for unseen classes. The number of human activities could be potentially unbounded. Therefore, a valuable aspect of autoencoding is generating features that effectively describe all activity categories, predefined and undefined, without retraining the autoencoder for each new class.  To verify this capability, we train an activity recognition model on the HAR training set by excluding all samples of one of the classes. After training, the entire set of training and test data are encoded by activity2vec. As shown in Table \ref{table:zeroshot}, the RF model achieved an acceptable performance for all left-out classes. We conclude that the activity2vec framework can produce adequate features for unseen classes if it trains on enough instances of related classes.

\begin{table}[t]
    \centering
    \caption{F1-score per class on HAR test set.}
    \resizebox{\columnwidth}{!}{
    \begin{tabular}{l c c c c c c} \toprule[2pt]
         & walking & walking up & walking down & sitting & standing & laying \\ \hline
        Raw features  & 0.84 & 0.81 & 0.89 & 0.79 & 0.82 & {\bf 1.00} \\ \hline
        Hand-crafted features & 0.93 & 0.89 & 0.90 & {\bf 0.90} & {\bf 0.91} & {\bf 1.00} \\ \hline
        Activity2Vec & {\bf 0.97} & {\bf 0.95} & {\bf 0.94} & 0.83 & 0.85 & {\bf 1.00} \\ \bottomrule[2pt] 
    \end{tabular}}
    \label{table:HAR-f1score}
\end{table}
\begin{table}[t]
    \centering
    \caption{Intra-class similarities for HAR dataset. Here, lower values are preferred as they indicate a greater cohesiveness of instances within each learned class.}
    \resizebox{\columnwidth}{!}{
    \begin{tabular}{l c c c c c c} \toprule[2pt]
         & walking & walking up & walking down & sitting & standing & laying \\ \hline
        Hand-crafted feature & 13.17 & 13.11 & 13.74 & 11.98 & 11.90 & 12.42 \\ \hline
        Activity2Vec & 4.02 & 4.21 & 4.04 & 1.85 & 1.15 & 2.60 \\ \bottomrule[2pt] 
    \end{tabular}}
    \label{table:HAR-intra}
\end{table}
\begin{table}[t]
    \centering
    \caption{F1-score per class using leave-one-class-out activity2vec encoding for the HAR test set.}
    \resizebox{\columnwidth}{!}{
    \begin{tabular}{l c c c c c c} \toprule[2pt]
        \textbf{Excluded class} & walking & walking up & walking down & sitting & standing & laying \\ \hline
        \textbf{walking}  & 0.95 & 0.95 & 0.95 & 0.84 & 0.85 & 1.00 \\ \hline
        \textbf{walking up}  & 0.96 & 0.93 & 0.92 & 0.84 & 0.84 & 1.00 \\ \hline
        \textbf{walking down}  & 0.94 & 0.96 & 0.89 & 0.83 & 0.86 & 1.00 \\ \hline
        \textbf{sitting} & 0.98 & 0.96 & 0.97 & 0.84 & 0.86 & 1.00 \\ \hline
        \textbf{standing}  & 0.97 & 0.94 & 0.94 & 0.82 & 0.85 & 1.00 \\ \hline
        \textbf{laying} & 0.97 & 0.94 & 0.95 & 0.84 & 0.86 & 0.98 \\ \bottomrule[2pt] 
    \end{tabular}}
    \label{table:zeroshot}
\end{table}

\begin{table*}[t]
    \centering
    \caption{F1-score per class on H101 test set.}
    \resizebox{\textwidth}{!}{
    \begin{tabular}{l c c c c c c c c c c c c} \toprule[2pt]
         & bath & bed to toilet & cook & eat & enter & leave & other activity & hygiene & relax & sleep & dishes & work  \\ \hline
        One-hot encoded features & 0.82 & 0.07 & 0.74 & 0.17 & 0.76 & 0.40 & \textbf{0.87} & 0.87 & \textbf{0.93} & \textbf{0.75} & \textbf{0.25} & 0.00\\ \hline
        Hand-crafted features & 0.82 & \textbf{0.55} & \textbf{0.75} & \textbf{0.27} & \textbf{0.85} & \textbf{0.42} & 0.85 & \textbf{0.89} & 0.92 & 0.74 & 0.24 & 0.00\\ \hline
        Activity2Vec & \textbf{0.83} & 0.07 & \textbf{0.75} & 0.19 & 0.71 & 0.33 & \textbf{0.87} & 0.87 & \textbf{0.93} & \textbf{0.75} & \textbf{0.25} & 0.00 \\ \bottomrule[2pt] 
    \end{tabular}}
    \label{table:HH101-f1score}
\end{table*}
\begin{table*}[t]
    \centering
    \caption{Intra-class similarities for HH101 dataset. Here, lower values are preferred as they indicate a greater cohesiveness of instances within each learned class.}
    \begin{tabular}{l c c c c c c c c c c c c} \toprule[2pt]
         & bath & bed to toilet & cook & eat & enter & leave & other activity & hygiene & relax & sleep & dishes & work  \\ \hline
        Hand-crafted feature & 0.97 & 1.00 & 1.82 & 1.96 & 1.81 & 1.89 & 2.71 & 2.56 & 2.18 & 2.10 & 2.06 & 1.27\\ \hline
        Activity2Vec & 1.34 & 1.74 & 1.93 & 2.22 & 2.17 & 2.12 & 2.40 & 2.23 & 2.01 & 2.28 & 2.02 & 1.8 \\ \bottomrule[2pt] 
    \end{tabular}
    \label{table:HH101-intra}
\end{table*}

Additionally, the activity2vec features exhibit a higher intra-class similarity than the handcrafted features as shown in Tables \ref{table:HAR-intra} and \ref{table:HH101-intra}. This indicates that the learned features are effective at grouping same-class instances together in the representational space.

\begin{figure}[h]
  \centering
  \subfloat[Handcrafted features.]{\includegraphics[width=0.25\textwidth]{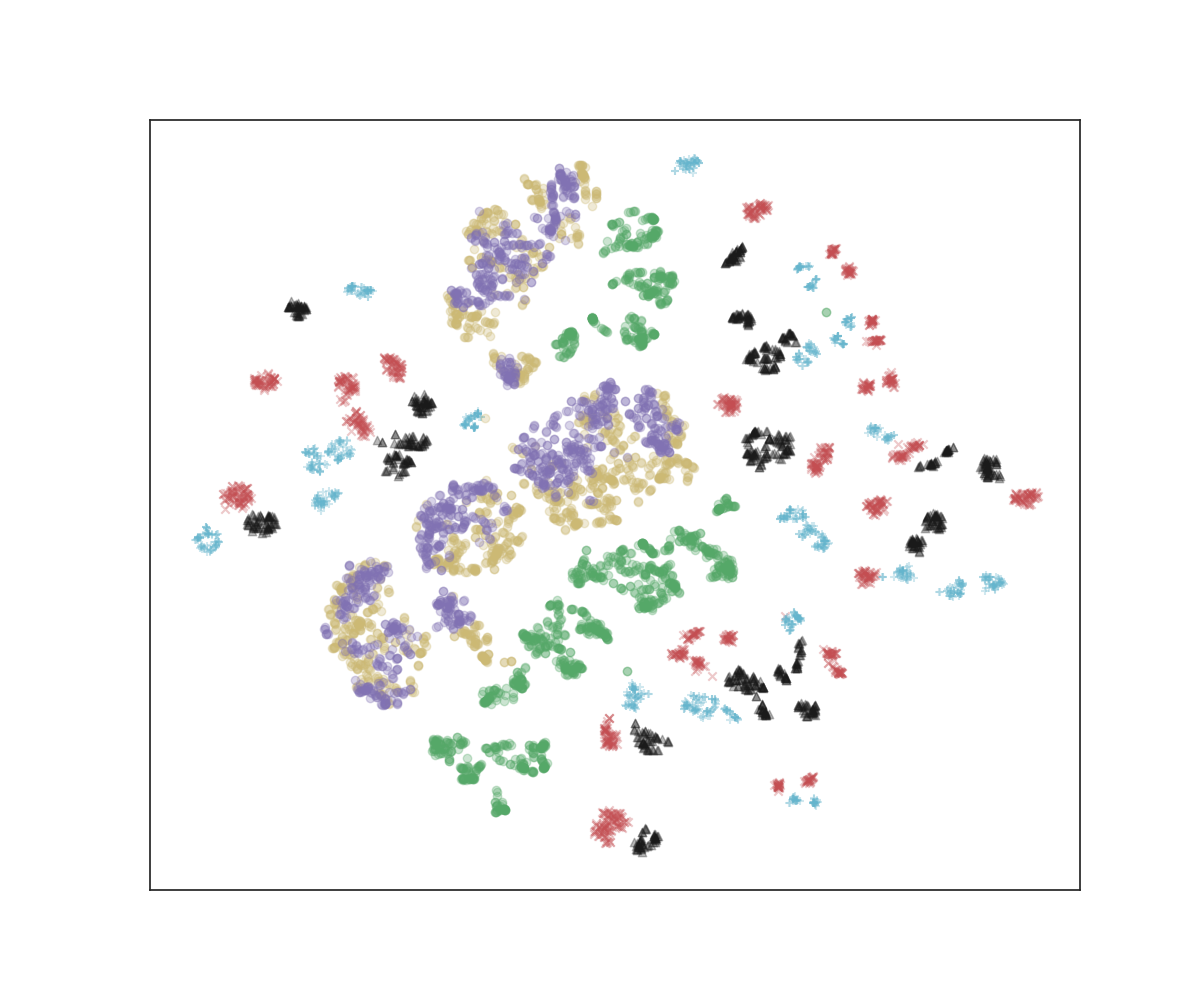}\label{fig:f1}}
%   \hfill
  \subfloat[Encoded features.]{\includegraphics[width=0.25\textwidth]{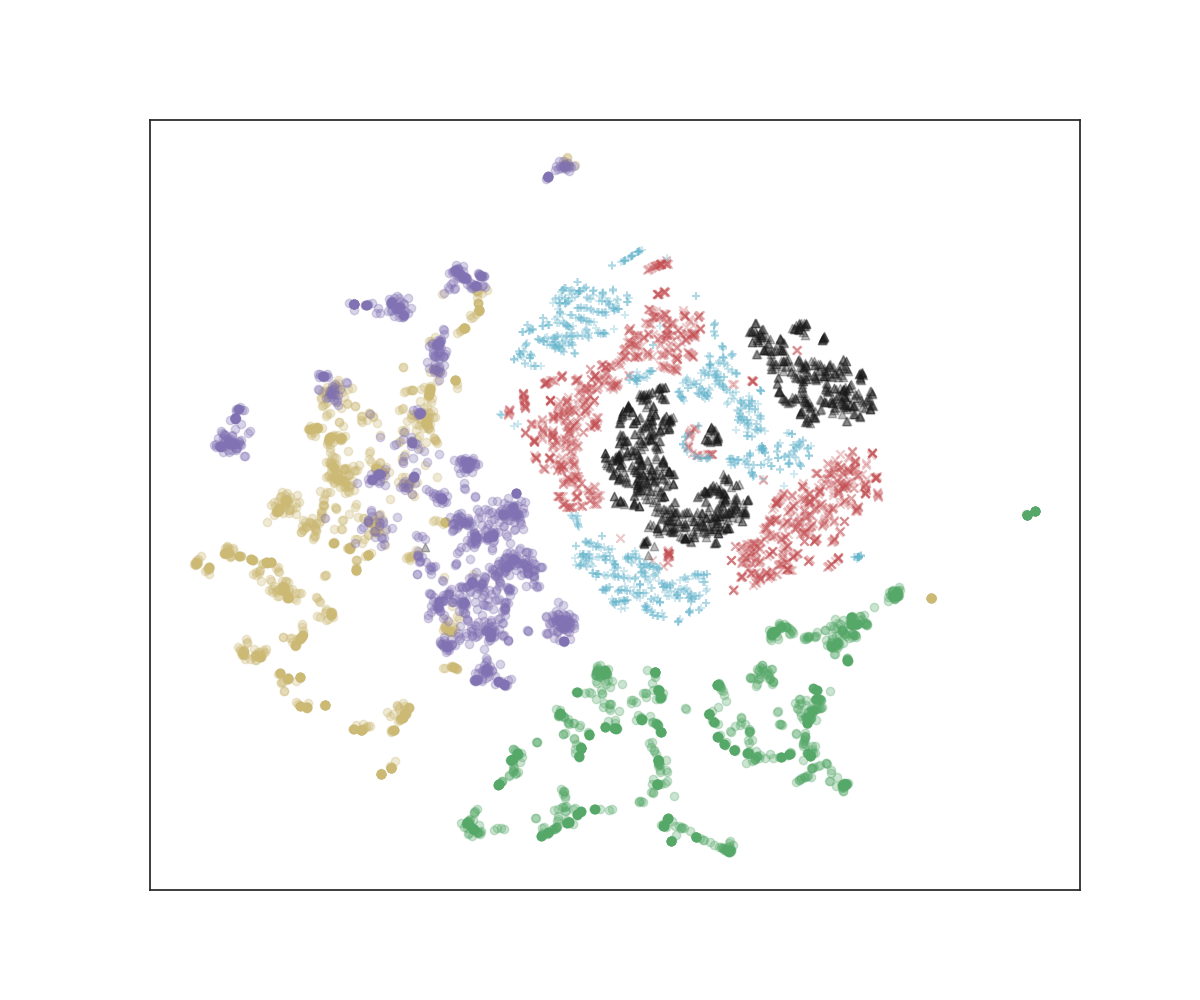}\label{fig:f2}}
  \caption{t-SNE projection of handcrafted features and encoded features via activity2vec on the HAR training set. (red=walking, blue=walking upstairs, black=walking downstairs, yellow=sitting, purple=standing, green=laying)}
  \label{figure:har-tsne}
\end{figure}

We visualize hand-crafted features and the activity2vec-generated sensor vector embeddings for HAR with t-SNE \cite{maaten2008visualizing}  in Figure \ref{figure:har-tsne}. Here, we observe that activity2vec features are in fact well separated versus the hand-crafted features based on their classes. We further observe that similar classes appear geometrically close together in this visualization.

\section{Conclusions and Future Work}

In this paper, we introduce activity2vec framework to extract meaningful features from sensor-based time series data. Activity2vec utilizes a RNN encoder-decoder framework to represent sensor data collected from body-worn sensors or ambient sensors. To best of our knowledge, this is the first work which is applied a seq2seq model to ADL activity data for extracting features. Activity2Vec achieves richer information than is available in raw sensor data or from hand-crafted features. Beyond utilizing learned features for classification, the embedding feature representations make it possible to ask questions about the similarity between unknown, unlabeled activities and known activities. Investigating the effectiveness and usefulness of this is a direction for our future work. Another avenue of work is understanding whether the learned features encode identifiable characteristics of a person and whether the evolution of the latent space corresponds to significant changes in behavior.

\begin{acks}
This work was supported in part by National Institutes of Nursing grant R01NR016732.
\end{acks}

\bibliographystyle{ACM-Reference-Format}
\bibliography{sample-base}

\end{document}